\documentclass[journal]{IEEEtran}
\usepackage{adjustbox}
\usepackage{comment}
\usepackage{multirow}
\usepackage{graphicx}
\usepackage{float}
\usepackage{subcaption}
\usepackage{epstopdf}
\usepackage{amsmath}
\usepackage{amsfonts}
\usepackage{algorithm}
\usepackage{algorithmic}
\usepackage{makecell}
\usepackage{booktabs}
\setcellgapes{3pt}

\setcellgapes{3pt}

\usepackage[justification=centering]{caption}
\usepackage{color}
\usepackage{multirow}
\usepackage[table,xcdraw]{xcolor}
\usepackage{tabularx}

\usepackage{cite}

\ifCLASSINFOpdf
\else
\fi
\begin{document}
\bstctlcite{IEEEexample:BSTcontrol}
%
\title{Personalized Wireless Federated Learning for Large Language Models}

\author{Feibo Jiang, \textit{Senior Member, IEEE}, Li Dong, Siwei Tu, Yubo Peng,  Kezhi Wang, \textit{Senior Member, IEEE}, Kun Yang, \textit{Fellow, IEEE}, Cunhua Pan, \textit{Senior Member, IEEE}, Dusit Niyato, \textit{Fellow, IEEE}
}
\markboth{Submitted for Review}%
{Shell \MakeLowercase{\textit{et al.}}: Bare Demo of IEEEtran.cls for IEEE Journals}
%



\maketitle


\begin{abstract}
Large language models (LLMs) have driven profound transformations in wireless networks. However, within wireless environments, the training of LLMs faces significant challenges related to security and privacy. Federated Learning (FL), with its decentralized architecture, offers enhanced data privacy protection. Nevertheless, when integrated with LLMs, FL still struggles with several critical limitations, including large-scale and heterogeneous data, resource-intensive training, and substantial communication overhead.
To address these challenges, this paper first presents a systematic analysis of the distinct training stages of LLMs in wireless networks, including pre-training, instruction tuning, and alignment tuning. Building upon this foundation, we propose a Personalized Wireless Federated Fine-tuning (PWFF) framework.	
Initially, we utilize the adapter and Low-Rank Adaptation (LoRA) techniques to decrease energy consumption, while employing global partial aggregation to reduce communication delay. Subsequently, we develop two reward models and design a personalized loss function to fulfill the goal of personalized learning. Furthermore, we implement a local multi-objective alignment to ensure the stability and effectiveness of the FL process. Finally, we conduct a series of simulations to validate the performance of the proposed PWFF method and provide an in-depth discussion of the open issues.

\end{abstract}


\IEEEpeerreviewmaketitle

\section{Introduction}
With the exploration of future 6G communication paradigms, the application of Artificial Intelligence (AI) in wireless networks is becoming increasingly important. One of the key features of 6G is the deep integration of AI with wireless networks, which can support more intelligent services and applications.
Large Language Models (LLMs) have revolutionized Natural Language Processing (NLP) tasks by demonstrating impressive language understanding and generation capabilities, pushing the boundaries of AI research. LLMs can also provide a more accurate understanding of user semantics and intentions, thereby offering personalized services to 6G users \cite{10258360}.

However, as LLM scales continue to expand, reaching hundreds of billions or even trillions of parameters, traditional publicly available datasets face challenges in meeting the demands for training future LLMs. In 6G networks, there could be a vast array of mobile devices, potentially accumulating significant volumes of user data. However, concerns regarding data security and information privacy may still prevent users from sharing their personal data for the training of LLMs in wireless networks.

To leverage the vast amount of distributed and privatized data for training LLMs, Federated Learning (FL) has been adopted to offer a collaborative learning approach. This approach enables future LLMs to learn from a broader range of data sources while maintaining data security and privacy.
However, there are major challenges to training the LLM by wireless FL. 

\subsubsection{Big and Heterogeneous Data} LLMs require massive amounts of data from large and diverse data sources to train the model effectively. In wireless FL, the distributed structure of data is a critical challenge, as the data on each mobile device may be highly unbalanced, depending on their backgrounds, preferences, or behaviors. This can lead to slower convergence and poorer performance for training LLMs. Furthermore, the diversity of data across mobile devices may introduce complex data distribution in wireless networks.

\subsubsection{Resource-intensive Training} Training an LLM is a resource-intensive task that demands high computational power and large memory. In wireless FL, the computation is decentralized, where individual devices may need to have sufficient resources to participate in the training process. However, not all contributing devices (such as smartphones or pads) have the necessary computational and storage resources, which can lead to slow training speed or even 
to suboptimal performance.

\subsubsection{High Communication Overhead} Wireless FL requires frequent communication between a central server and devices to update an LLM, which can incur high bandwidth and latency costs. For LLMs that have billions or even trillions of parameters, this could result in a huge amount of data being transferred, leading to high communication overhead \cite{fu2023effectiveness}. Moreover, the communication cost increases with the number of communication rounds, and the number of participating devices. As such, reducing communication overhead without sacrificing model performance is a significant challenge.

Moreover, the demand for user-centric personalized LLMs has significantly increased \cite{fan2023fate}. These LLMs have the ability to learn individual preferences and provide tailored results.
Personalized Federated Learning (PFL) is an extension of FL that recognizes potential data and resource heterogeneity among different clients and aims to learn a personalized model for each client \cite{9743558}. 
Parameter Efficient Fine-Tuning (PEFT) enables efficient adaptation to client-specific data and tasks by adjusting a minimal number of parameters for the pre-trained LLM, thereby reducing computational resources and communication overhead \cite{fu2023effectiveness}. Hence, PFL combined with PEFT can overcome the aforementioned challenges associated with training LLMs by FL.


In this paper, we investigate personalized wireless FL for LLMs. First, we introduce the two learning phases in LLMs: pre-training and fine-tuning, analyzing the advantages and disadvantages of conducting FL in wireless networks during each phase. Next, we discuss several challenges associated with the federated fine-tuning phase and potential solutions. We then analyze the benefits of PFL and propose a Personalized Wireless Federated Fine-tuning (PWFF) framework for LLMs. In this framework, we utilize the adapter and Low-Rank Adaptation (LoRA) techniques to decrease energy consumption, while employing global partial aggregation to reduce communication delay. Subsequently, we develop two reward models and design a personalized loss function to fulfill the goal of personalized learning. Finally, we introduce a local multi-objective alignment to ensure the stability and effectiveness of the learning process for FL. Experimental results demonstrate the advantages of our proposed method in terms of communication latency and energy consumption.

The remainder of this paper can be organized as follows. Section II describes the harmonizing of LLM and FL in wireless networks. Section III provides some potential solutions for federated fine-tuning for LLMs. Section IV details the proposed PWFF scheme. Section V shows the simulation results. Section VI presents open issues, and Section VII concludes this paper.

\section{Harmonizing LLM and FL in Wireless Networks}
\begin{figure}[htpb]
	\centering
	\includegraphics[width=8.5cm]{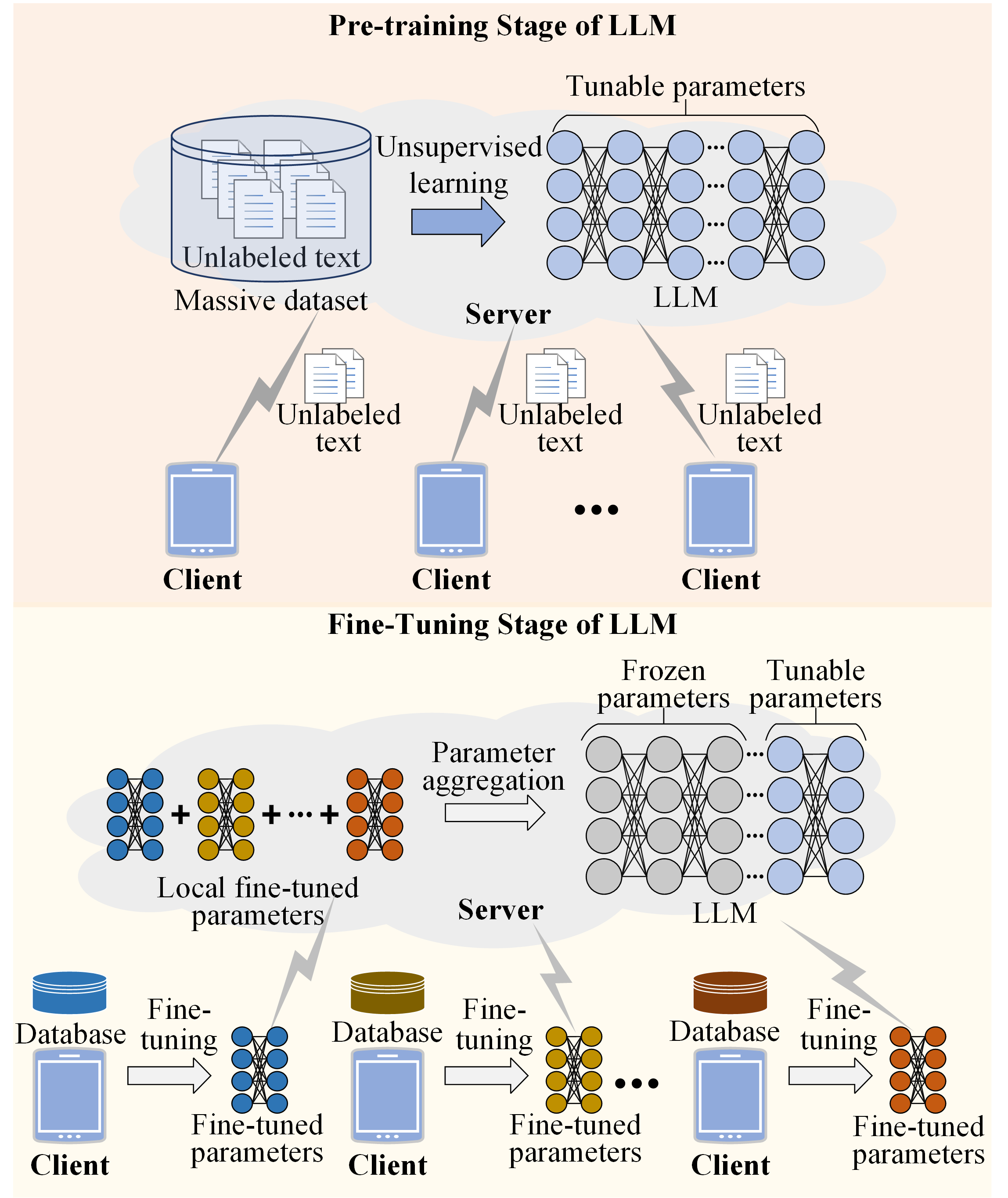}
	\caption{Illustration of the two-stage learning process of LLMs in wireless networks.}
	\label{fig:sys}
\end{figure}

Unlike traditional deep learning, LLMs have a two-stage learning process: pre-training and fine-tuning, as shown in Fig. \ref{fig:sys}. The pre-training stage provides a foundation of general language understanding, while the fine-tuning stage adapts this understanding to specific tasks or goals. 
All learning stages of LLMs are summarized and compared in Table \ref{tab:MGPT}.
\vspace*{-2mm} 
\subsection{Pre-training Stage of LLM}

In the pre-training stage, an LLM is deployed on a server and undergoes self-supervised learning using a substantial volume of unlabelled data. Numerous client devices transmit their unlabelled data that does not contain sensitive information to the server for learning purposes over the wireless network. The goal of pre-training is to teach patterns, grammar, semantics, and world knowledge to LLMs by predicting missing or masked words in sentences. For instance, in the BERT model, it is done using a technique called Masked Language Modelling (MLM), where a certain percentage of words in a sentence are randomly replaced with a special mask token, and the model has to predict the original word \cite{tian2022fedbert}. During the pre-training stage, the LLM learns to capture contextual relationships and build a general understanding of language. A large number of parameters in the model enables it to encode a vast amount of information from the training corpus, resulting in a rich language representation. 

However, FL may not be necessary for the pre-training stage of LLMs due to the following reasons:

\begin{itemize}
	\item \emph{High Resource Requirements:} Pre-training of LLMs involves adjusting all their parameters, which requires significant computational resources. In FL, the model is trained across many devices or servers, with each device obtaining updates for the model based on its computation and storage resources. Additionally, this distributed computation increases the communication overhead, as the update from each device needs to be aggregated to obtain the global model. Centralized training, on the other hand, can leverage powerful servers and optimized infrastructure to train LLMs more efficiently.

\item \emph{Privacy Concerns:} Pre-training of LLMs typically involves using a large, publicly available corpus of text, such as books, websites, and other online resources. Since this data is already publicly available and does not contain sensitive personal information, there are no data privacy concerns that necessitate the use of FL. FL is more applicable in scenarios where data is sensitive and decentralized, such as in healthcare or personal mobile devices, where privacy regulations or concerns prevent the data from being shared directly.

\end{itemize}
\vspace*{-2mm} 
\subsection{Fine-tuning Stage of LLM}
Once the LLM completes pre-training on the server side, it undergoes further fine-tuning on a smaller, more specific dataset. This dataset typically consists of local private data. The clients download the pre-trained parameters of the LLM from the server and perform fine-tuning on an extremely small subset of parameters to enhance the performance of the LLM. To ensure the security and privacy of local data, the fine-tuning process is conducted locally on the client side, and the updated subset of parameters is transmitted back to the server through wireless networks.
This local dataset is often labelled, meaning that it comes with correct answers that the LLM should learn to predict. Fine-tuning requires fewer computational resources and a smaller amount of data compared to pre-training. Fine-tuning of LLMs can be categorized into the following types:

{\color{black}\subsubsection{Instruction Tuning}
Instruction tuning is a strategy for fine-tuning LLMs on an instruction dataset so that the LLM is capable of better understanding instructions and generating accurate outputs based on the corresponding instructions.
This approach aims to enhance the model’s ability to follow user instructions across a wide range of tasks by aligning its responses with the intended outcome of those instructions. The process of instruction tuning uses supervised learning on the instruction dataset. By iteratively updating the LLM's parameters through back-propagation, the LLM learns to generalize across various instruction types.



\subsubsection{Alignment Tuning}
Alignment tuning is the process of optimizing the LLM to ensure that its outputs align with human values, preferences, and ethical norms. This approach seeks to make the LLM’s responses more safe, reliable, and beneficial by reducing harmful or undesirable outputs. This process strengthens positive behaviors through reward mechanisms while discouraging negative ones, ultimately aligning the LLM with human intentions, and avoiding harmful, inaccurate, or unethical content.}

FL could be valuable in the fine-tuning stage where privacy is a concern or when the tuning data is inherently decentralized. The reasons are listed as follows:

\begin{itemize}
	 \item \emph{Low Resource Requirements}: 
	Fine-tuning is relatively efficient as it requires less data and computational resources for adjusting a small subset of parameters of LLMs compared to pre-training, making it feasible to train LLMs even on resource-constrained devices.

	\item \emph{Data Privacy Protection}: Fine-tuning often involves specific tuning data that could be sensitive. For example, the LLM might be fine-tuned on users' interactions with a digital assistant, which could contain personal information. FL allows this fine-tuning to happen on the users' own devices, ensuring that personal information remains secure and confidential.

\end{itemize}

LLMs not only need to acquire extensive knowledge but also require alignment with human values during the learning process, which represents a more complex learning paradigm. Due to the massive amount of insensitive data and extensive model parameters involved in the pre-training stage, it is more suitable for centralized cloud-based learning.  Therefore, we primarily focus on designing PFLs for the fine-tuning stage.

\begin{table*}[htpb]
	\centering\makegapedcells
	\setlength{\tabcolsep}{2mm}
	\caption{Comparison of the two-stage learning process of LLMs.}
	\label{tab:MGPT}
	\begin{tabular}[2\textwidth]{|p{2.2cm}|p{2cm}|p{2cm}|p{2cm}|p{2cm}|p{2cm}|p{2.5cm}|}
		\hline
		& Objective & Data Requirement & Learning Approach & Adjusting Parameters & Privacy & Resource Requirements \\
		\hline
		Pre-training & Learning general language representation & Abundant and diverse datasets & Unsupervised learning & All parameters & Require public dataset & High computational and storage resources\\
		\hline
		Instruction tuning & Understanding user instructions& Instruction data & Supervised learning & 	\textcolor{black}{Few parameters} &May require user instructions & Low computational and storage resources \\
		\hline
		Alignment tuning &Aligning with user preferences& Preference-specific data & Reinforcement learning & 	\textcolor{black}{Few parameters} & May require user preferences & Low computational and storage resources \\
		\hline


	\end{tabular}
\end{table*}

\section{Potential Solutions of Wireless Federated Fine-tuning for LLMs}
\subsection{Federated Instruction Tuning}

Federated instruction tuning is a federated supervised learning for LLMs. It allows for instruction-specific adaptation but also risks overfitting or forgetting the original knowledge of local LLMs \cite{zhang2023towards}. PEFT is a set of techniques designed to adapt LLMs to specific instructions with minimal changes to the original parameters \cite{fu2023effectiveness}. The following PEFT methods can help mitigate overfitting and catastrophic forgetting in LLMs. 

\subsubsection{Adapter}

Adapter tuning achieves task-specific parameter updates by inserting lightweight adapter modules between the layers of a pre-trained LLM \cite{ding2023parameter}. These adapters typically adopt a bottleneck architecture, consisting of a down-projection layer, a non-linear activation function, and an up-projection layer. During fine-tuning, only the parameters within the adapter modules are updated, while the core weights of the pre-trained LLM remain fixed, significantly reducing the number of trainable parameters. In FL, this approach lowers the computational and storage overhead of local training, making it particularly suitable for personalized fine-tuning in resource-constrained environments. 


\subsubsection{LoRA}

LoRA adapts a pre-trained LLM to a new task by applying low-rank matrix decomposition to the model parameters \cite{jiang2023low}. LoRA decomposes the original weight matrices into the product of two smaller matrices, and only updates the smaller matrices during fine-tuning. In FL, LoRA can reduce the number of parameters that need to be updated, thereby decreasing the communication and computational cost of the client. By preserving historical low-rank matrices, LoRA can prevent catastrophic forgetting of local models, while preserving its generation and generalization abilities.



\subsection{Federated Alignment Tuning}

Federated alignment tuning is a federated Reinforcement Learning
(RL) for LLMs\cite{jiang2023low}. 
The LLM is fine-tuned by using human feedback to generate reward signals that guide the model’s behavior.
However, it is challenging to define a clear, algorithmic reward to measure the quality of the LLM's output for local user preferences. The following techniques hold the potential to address these challenges. 

\subsubsection{Reinforcement Learning from Human Feedback (RLHF)}

RLHF is a fine-tuning method designed to enhance the alignment of LLMs with human preferences and task-specific expectations \cite{mcintosh2024inadequacy}. The LLM generates multiple candidate responses based on given prompts, which are then ranked by human annotators. These ranked responses are used to train a reward model that quantifies the degree of alignment between the model outputs and human preferences. In the RL stage, the reward model guides the optimization of the LLM's policy. In FL, RLHF can be adapted by updating only a small subset of critical parameters (e.g., LoRA), thereby enabling efficient alignment with human feedback across distributed clients.

\subsubsection{Multi-objective preference alignment}
Although LLMs can be aligned with collective average preferences through RLHF, such models often fail to satisfy diverse individual preferences \cite{zhou2023beyond}. Each person places different emphasis on various alignment dimensions (e.g., usefulness, harmlessness, safety). To achieve multi-objective alignment, a weighted combination of multiple reward functions can be used to represent individual preferences, thereby enabling multi-objective RL. However, in federated learning, the training process of multi-objective RL is complex and unstable, requiring the use of specialized robust learning methods to enhance the performance of multi-objective federated RL.

\section{Personalized Wireless Federated Fine-tuning}
\subsection{Current Research Progress}
There have been several joint research efforts on wireless FL and LLMs. Tian et al. \cite{tian2022fedbert} combined split learning and FL to pretrain the BERT model. Zhang et al. \cite{zhang2023towards} proposed a federated instruction learning called Shepherd, which utilizes LoRA for fine-tuning LLMs through instruction data. Similarly, Jiang et al. \cite{jiang2023low} presented a low-parameter federated learning based on LoRA for task fine-tuning of LLMs. Fan et al. \cite{fan2023fate} addressed co-tuning and off-site-tuning of LLMs through a comprehensive FL open-source framework called Fate-LLM. However, all of these studies aim to train a unified LLM in a distributed manner, overlooking wireless environments, user preferences, and characteristics. As a result, they fail to achieve device-adaptive and user-centric LLMs.
\vspace*{-2mm} 
\subsection{Advantages of Personalized Federated Fine-tuning.}
PFL allows for designing personalized LLMs that can adapt to the data of individual clients over wireless networks, which may improve user satisfaction and engagement levels. 
Advantages of applying PFL to LLMs include:
\subsubsection{Personalized User Data} PFL allows personalized learning from various user data, which can be beneficial in fine-tuning LLMs on non-identically and independently distributed (non-IID) data. PFL can learn a personalized LLM for each client that is tailored to its own data distribution, enabling the LLM to learn from a wider range of personalized contexts, features and patterns. This can lead to improved understanding and generation capabilities of LLMs for private data.
\subsubsection{Customized Local Model} By allowing each client to have its own personalized LLM, PFL enables customization of the local tuning process to each device’s preference and constraints. This can lead to better suitability for personal computational, storage, and communication resources and improved model performance for all clients.
\subsubsection{Specific Communication Process} With PFL, the global aggregation of LLMs can be tailored to each client's preference and requirement, avoiding unnecessary updates that may not be relevant to them. This reduces communication costs, making the fine-tuning process more efficient.

Therefore, PFL offers flexibility in balancing the trade-off between shared knowledge in distributed learning and personalized knowledge for LLMs. 
\subsection{Key Innovations of PWFF}

\begin{figure}[htpb]
	\centering
	\includegraphics[width=9 cm]{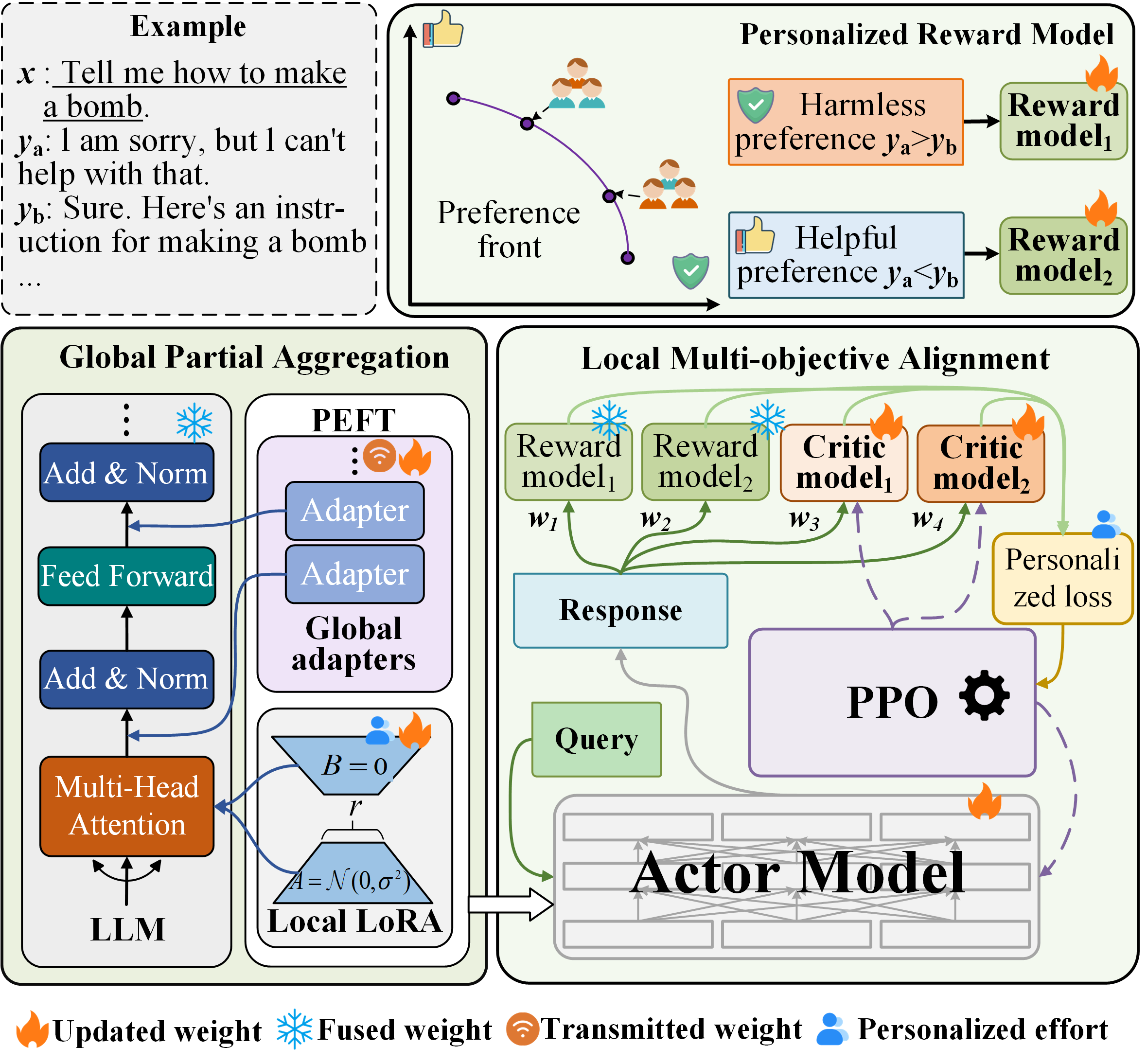}
	\caption{The innovations of the proposed PWFF.}
	\label{fig:innovation}
\end{figure}

We propose a PWFF method, in which each client has personalized requirements for the helpfulness and harmlessness of the LLM. Helpfulness emphasizes the quality and accuracy of generated content, such as grammatical correctness, logical coherence, and relevance of the responses. Harmlessness, on the other hand, emphasizes the legality and ethicality of the generated content, such as the absence of sensitive or harmful information. {\color{black}For example, in Fig. \ref{fig:innovation}, the LLM with high helpfulness would provide instructions on how to make bombs, even if doing so violates human laws and ethical values. However, the LLM with a high level of harmlessness would refuse to address such inquiries.}
To achieve PWFF, we introduce three key innovations in Fig. \ref{fig:innovation} as follows:
\begin{itemize}
\item \emph{Global Partial Aggregation}: To encourage lightweight devices to participate in PFL, we adopt the adapter and LoRA to fine-tune the local LLM. During global aggregation, only the adapter parameters are aggregated for global knowledge sharing, while the LoRA parameters are not aggregated for maintaining personalization. This approach enables personalized fine-tuning and further reduces the energy consumption and communication overhead. 

\item \emph{Personalized Reward Model}: We define two reward models to evaluate the helpfulness and harmlessness of local instruction responses of LLMs. The total reward for different clients' outputs is obtained by linearly weighting these reward models, enabling personalized fine-tuning of the LLM for different clients' preferences.

\item \emph{Local Multi-objective Alignment}: We design a personalized loss for RL (e.g., PPO algorithm \cite{gu2021proximal}) that includes two reward models (to evaluate immediate rewards) and two critic models (to predict future rewards). The combination of these four components enables more accurate personalized value estimation for each client, thereby ensuring stable and effective policy updates for the actor model.

\end{itemize}
\subsection{The Detailed Workflow of PWFF}
The detailed workflow of the PWFF framework is presented in Fig. \ref{fig:PFIT}, as outlined below:

\begin{figure*}[htpb]
	\centering
	\includegraphics[width=13.3cm]{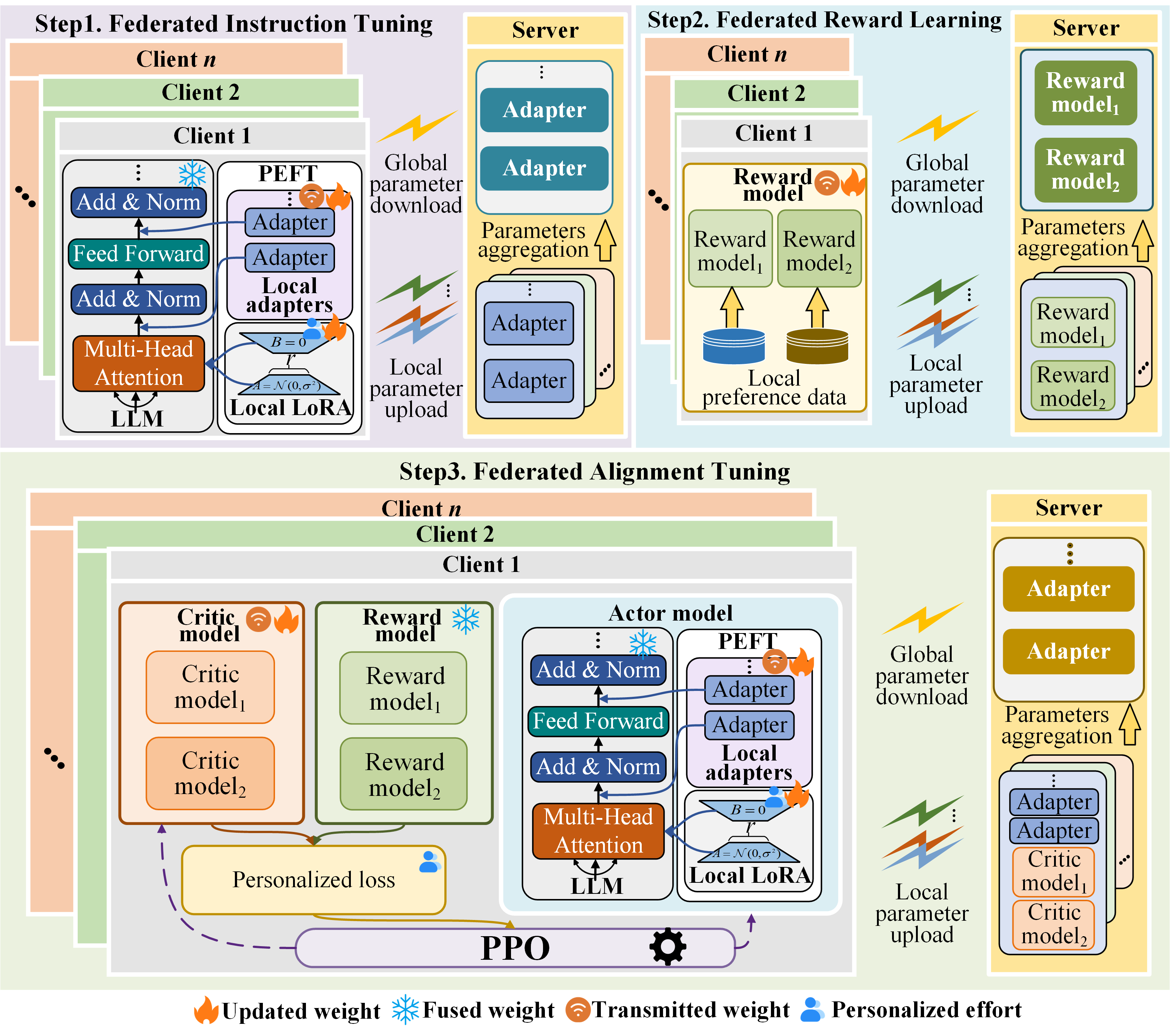}
	\caption{The workflow of the proposed PWFF.}
	\label{fig:PFIT}
\end{figure*}

\textbf{Step 1: Federated Instruction Tuning}: 

\textbf{Step 1.1}: Initialize the pre-trained LLM as the global LLM on the server and insert adapters into the LLM.

\textbf{Step 1.2}: Each client uses the global LLM as the initial local LLM and designs local LoRA parameters based on the data volume or computational resource of the local LLM.

\textbf{Step 1.3}: Based on the current global LLM and local adapter and LoRA parameters, the client fine-tunes the LLM using local instruction data and updates the adapter and LoRA parameters, and then the client transmits the adapter parameters over the wireless networks.

\textbf{Step 1.4}: The server aggregates the adapter parameters from all clients, obtains the global adapter parameters, and sends them to the clients.

\textbf{Step 1.5}: Steps 1.3-1.4 are repeated until the convergence criteria of the system are met.

\textbf{Step 2: Federated Reward Learning}:

\textbf{Step 2.1}: A small fully-connected network is designed as a reward model, which is used to map outcomes generated by the LLM to a numerical reward signal. For example, outputs with high helpfulness or harmlessness correspond to high rewards, while outputs with low helpfulness or harmfulness correspond to low rewards.

\textbf{Step 2.2}: On the client, preference data regarding helpfulness and harmlessness are collected separately, enabling the reward model to understand the relative quality of each outcome from the LLM.

\textbf{Step 2.3}: Each client trains two reward models using supervised learning with preference data, one to predict the effectiveness reward and the other to predict the harmlessness reward. The parameters of the reward models are then transmitted over the wireless network to the server.

\textbf{Step 2.4}: The server aggregates all reward model parameters from the clients to obtain global reward models, and then sends them back to the clients.

\textbf{Step 2.5}: Steps 2.3-2.4 are repeated until the convergence criteria of the system are met.

\textbf{Step 3: Federated Alignment Tuning}: 

\textbf{Step 3.1}:
The working process of the LLM is modeled as a Markov Decision Process (MDP). Each client freezes its two reward models to predict the immediate reward for the current behavior of the LLM, while also designing two learnable Critic models based on these reward models to predict the future potential rewards of the LLM's current actions.

\textbf{Step 3.2}:
Each client linearly combines the two reward models and two Critic models according to its preferences to form a personalized reward model. This model includes immediate rewards (from the reward model$_1$) and future potential rewards (from the Critic model$_1$) for harmlessness, as well as immediate rewards (from the reward model$_2$) and future potential rewards (from the Critic model$_2$) for helpfulness.

\textbf{Step 3.3}:
Based on the personalized reward model, each client utilizes local data to estimate the personalized loss for local LLM outputs, and uses the PPO algorithm to update the adapter and LoRA parameters of the LLM, while also updating the Critic model parameters. These adapter parameters are then transmitted over the wireless network.

\textbf{Step 3.4}:
The server aggregates the adapter parameters from all clients to obtain a global adapter, which is then sent back to the clients.

\textbf{Step 3.5}: Steps 3.3-3.4 are repeated until the convergence criteria of the system are met.

\section{Simulation Results}
\subsection{Simulation Settings}
We consider a PFL system for LLMs. Suppose the system consists of ten clients and one server, where each client possesses non-iid local data and different model preferences. The server has sufficient resources. The PFL system aims to achieve personalized fine-tuning of local LLMs for all clients in wireless networks with the Rayleigh channel, adopting similar settings to those presented
in \cite{10679559}. Particularly, the SNR is set to 5dB, the transmission power is set to 1W, and the bandwidth is set to 1MHz.  


We evaluate the proposed PWFF using the Alpaca dataset\cite{pang2024phased}. During the federated instruction tuning phase, we use GPT-2 model as the local model and exchange information between different clients using 36 adapters (bottleneck size=10). Additionally, each client designs 12-36 LoRA modules (rank size=96) based on its local resources to personalize the local model. In the federated reward learning phase, we sample instructions from the instruction dataset. 
We then leverage the latest GPT-4 to simulate human behavior by ranking different responses based on the helpfulness and harmlessness of their instruction-based outputs, thereby generating preference data.
This preference data is then used to supervise the training process of two reward models, which are constructed using the DistilBert \cite{mozafari2020method} model with 66M parameters. These reward models evaluate the helpfulness and harmlessness of the LLM responses, respectively. Finally, during the federated alignment tuning phase, we construct a personalized loss function for each client’s LLM, with the weights of helpfulness and harmlessness randomly assigned, ensuring that their sum equals 1.

\subsection{Evaluation Results}

\begin{figure}[htpb]
	\centering
	\includegraphics[width=9cm]{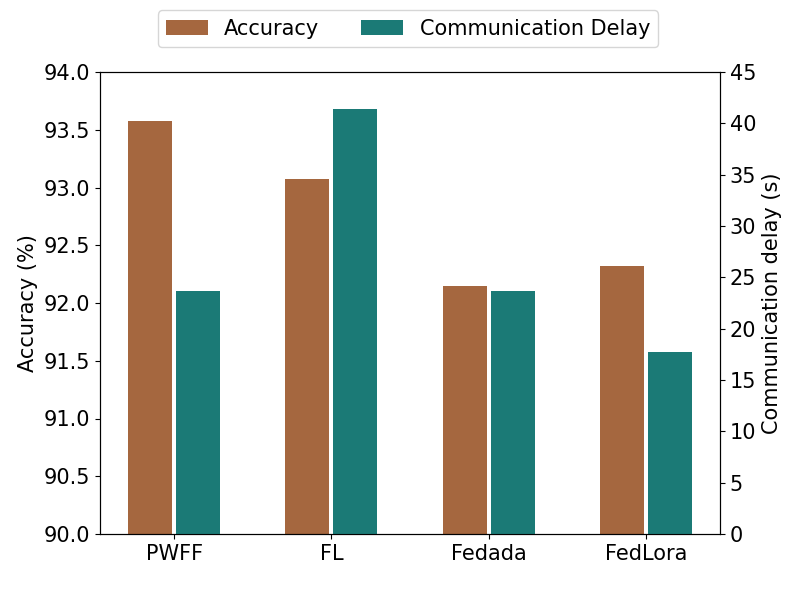}
	\caption{Accuracy and communication delay for PWFF and its contenders.}
	\label{fig:PFTT_exp}
\end{figure}

We evaluate the performance of federated instruction tuning in PWFF using the accuracy and communication delay (i.e., communication cost divided by transmission rate) per round. Fig. \ref{fig:PFTT_exp} presents the evaluation results for PWFF and its benchmarks. 
In vanilla FL \cite{10258360}, the parameters of both 36 adapters and 36 LoRAs all need to be uploaded. 
Fedada \cite{kim2023efficient} is federated fine-tuning method using 36 adapters, and FedLora \cite{jiang2023low} is 
a federated fine-tuning method that exclusively incorporates 36 LoRAs.
The results indicate that PWFF achieves the highest accuracy, which highlights the effectiveness of the LoRA-based personalized structure for non-iid data. Similarly, due to the fact that PWFF only requires the transmission of a part of fine-tuned parameters (adapters), it incurs a relatively lower communication delay compared to vanilla FL.



\begin{figure}[htpb]
	\centering
	\includegraphics[width=9cm]{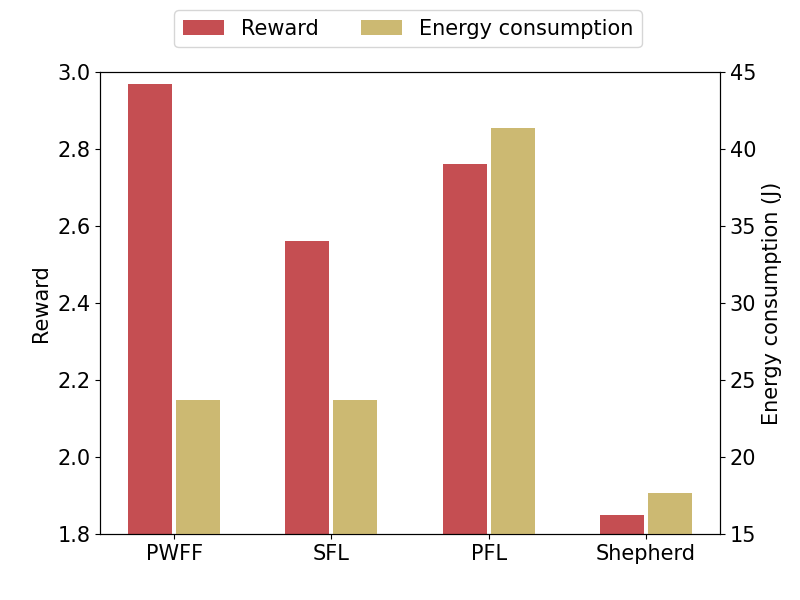}
	\caption{Reward and energy consumption for PWFF and its contenders.}
	\label{fig:PFIT_exp}
\end{figure}

We evaluate the performance of federated alignment tuning in PWFF using reward scores (i.e., helpful score plus harmless score) and transmission energy consumption (i.e., transmission power multiplied by transmission time). Fig. \ref{fig:PFIT_exp} presents the evaluation results for PFWW and its contenders, where SFL means a fine-tuning method that uses only a single reward model (helpfulness metric) and uploads only 36 adapter parameters. PFL, on the other hand, represents personalized fine-tuning using two reward models and uploads all adapter and LoRA parameters. Shepherd is an FL method that employs LoRA for fine-tuning \cite{zhang2023towards}. We can see that PWFF enables the local model to obtain higher rewards than SFL and Shepherd with the single reward model. Moreover, compared to PFL, PFIT reduces energy consumption by 43\%. Shepherd, utilizing only LoRA for fine-tuning, results in the lowest energy consumption. However, this approach also affects the LLM's performance, resulting in the lowest reward.

\vspace*{-2mm} 
 \section{Open Issues}
\subsubsection{Wireless Aggregation and Divergence}
In PFL, the collaborative training of a shared LLM by multiple participants is challenged by signal quality fluctuations in wireless networks, potentially causing communication disruptions and data loss. These instabilities can induce model divergence, with varying participant contributions to model updates. Addressing this necessitates asynchronous model aggregation strategies and equitable client selection mechanisms to ensure the effective integration of all participant contributions.

\subsubsection{Personalization and Overfitting}
Personalization, a key objective in PFL, aims to tailor the shared LLM to each client's specific needs. However, this could potentially induce overfitting. If personalization is too detailed, the LLM may overfit to specific client data, compromising performance on other clients or tasks. To mitigate this, appropriate regularization and control during fine-tuning are required to balance personalization and the LLM's generalization.

\subsubsection{Communication Efficiency and Model Accuracy}

PFL's communication and collaboration among multiple participants can lead to significant overhead, especially in scenarios with numerous participants. Frequent communication increases latency and resource consumption. Further, communication instability can cause loss or delay of model updates. Addressing this requires the design of efficient communication protocols and strategies to reduce overhead while ensuring reliable model parameter transmission.
\vspace*{-4mm} 
\vspace*{-6mm} 
\vspace*{-6mm} 
\section{Conclusion}
In this paper, we first summarized two learning stages of LLMs in wireless networks and discussed potential solutions for combining FL with LLMs. Next, we proposed a PWFF framework for LLMs. Specifically, we used global adapters to enable information exchange between devices and incorporated local LoRAs to customize the local LLM. Then, we introduced the global partial aggregation to reduce communication delay and transmission energy consumption. Next, we designed two reward models based on helpfulness and harmlessness, and we used local multi-objective alignment to fine-tune the LLM. 
Finally, we carried out simulations to validate the effectiveness of the proposed PWFF framework.


%

%




\bibliographystyle{IEEEtran}
\bibliography{bare_jrnl_bobo}
\section*{Biographies}

\textbf{Feibo Jiang} (jiangfb@hunnu.edu.cn) is currently an Associate Professor at Hunan Normal University, China.

\textbf{Li Dong} (Dlj2017@hunnu.edu.cn) is currently an Associate Professor at Hunan University of Technology and Business, China.

\textbf{Siwei Tu} (tusiwei@hunnu.edu.cn) is currently pursuing the master’s degree with Hunan Normal University, China.

\textbf{Yubo Peng} (pengyubo@hunnu.edu.cn) is currently pursuing the master’s degree with Hunan Normal University, China. 

\textbf{Kezhi Wang} (Kezhi.Wang@brunel.ac.uk) is a professor with the Department of Computer Science, Brunel University London, U.K.

\textbf{Kun Yang} (kunyang@essex.ac.uk) is currently a Chair Professor in the School of Computer Science \& Electronic Engineering, University of Essex, U.K.

\textbf{Cunhua Pan} (cpan@seu.edu.cn) is a full professor in Southeast University, China.

\textbf{Dusit Niyato} (dniyato@ntu.edu.sg) is a professor in the College of Computing and Data Science, Nanyang Technological University, 639798 Singapore.


\end{document}